\documentclass[11pt]{article}
\usepackage[hyperref]{emnlp-ijcnlp-2019}
\usepackage{times}
\usepackage{latexsym}
\usepackage{amsmath}
\usepackage{graphicx}
\usepackage{subcaption}
\usepackage[ruled]{algorithm2e}
\usepackage{paralist}
\usepackage{amsfonts}

\usepackage{url}

\newcommand{\Sref}[1]{\S\ref{#1}}

\newcommand{\Fref}[1]{Figure~\ref{#1}}
\newcommand{\tref}[1]{table~\ref{#1}}
\newcommand{\Tref}[1]{Table~\ref{#1}}

\newcommand{\ignore}[1]{}

\newcommand{\yt}[1]{\textcolor{blue}{\bf\small [#1 --YT]}}
\newcommand{\sw}[1]{\textcolor{magenta}{\bf\small [#1 --SW]}}
\newcommand{\sk}[1]{\textcolor{cyan}{\bf\small [#1 --SK]}}
\newcommand{\nascomment}[1]{\textcolor{olive}{\bf\small [#1 --NAS]}}

 \aclfinalcopy 

\title{Topics to Avoid: 
Demoting Latent Confounds in Text Classification} 

\author{Sachin Kumar$^\diamondsuit$ \quad Shuly Wintner$^\clubsuit$ \quad Noah A. Smith$^{\heartsuit\spadesuit}$ \quad Yulia Tsvetkov$^\diamondsuit$ \\
$^\diamondsuit$Language Technologies Institute, Carnegie Mellon University, Pittsburgh, PA, USA \\
$^\clubsuit$Department of Computer Science, University of Haifa, Haifa, Israel \\
$^\heartsuit$Paul G. Allen School of Computer Science \& Engineering, \\ University of Washington, Seattle, WA, USA \\
$^\spadesuit$Allen Institute for Artificial Intelligence, Seattle, WA, USA\\
\texttt{\small sachink@cs.cmu.edu, shuly@cs.haifa.ac.il}\\\texttt{\small  nasmith@cs.washington.edu, ytsvetko@cs.cmu.edu}}
\date{}

\begin{document}
\maketitle

\begin{abstract}
  Despite impressive performance on many text classification tasks, deep neural networks tend to learn frequent superficial patterns that are specific to the training data and do not always generalize well. In this work, we observe this limitation with respect to the task of \emph{native language identification}. We find that standard text classifiers which perform well on the test set end up learning topical features which are confounds of the prediction task (e.g., if the input text mentions Sweden, the classifier predicts that the author's native language is Swedish). 
  We propose a method that represents the latent topical confounds and a model which ``unlearns'' confounding features by predicting both the label of the input text and the confound; but we train the two predictors adversarially in an alternating fashion to learn a text representation that predicts the correct label but is less prone to using information about the confound. We show that this model generalizes better and learns features that are indicative of the writing style rather than the content.\footnote{The code is available at: \url{https://github.com/Sachin19/adversarial-classify}}

\end{abstract}


\section{Introduction}


Text classification systems based on neural networks are biased towards learning frequent spurious correlations in the training data that may be confounds in the actual classification task \citep{leino2018biasamplify}.
A major challenge in building such systems is to discover features that are not just correlated with the signals in the training data, but are true indicators of these signals, and therefore generalize well. 

For example, \newcite{kiritchenko2018examining} found that sentiment analysis systems implicitly overfit to demographic confounds, systematically amplifying the intensity ratings of posts written by women.
\newcite{zhao2017men} showed that visual semantic role labeling models implicitly capture actions stereotypically associated with men or women (e.g., \textit{women are cooking} and \textit{men are fixing a faucet}), and in cases of higher model uncertainty assign stereotypical labels to actions and objects, thereby amplifying social biases found in the training data.  

We focus on the task of \emph{native language identification} (L1ID), which aims at automatically identifying the native language (L1) of an individual based on their language production in a second language (L2, English in this work). The aim of this task is to discover stylistic features present in the input that are indicative of the author's L1. However, a model trained to predict L1 is likely to predict that a person is, say, a native Greek speaker, if the texts authored by that person mention Greece, because the training data exhibits such topical correlations (\Sref{sec:motivation-to-demote}). 

This problem is the focus of our work, and we address it in two steps. First, we introduce a novel method for representing \emph{latent} confounds. 
Recent relevant work in the area of domain adaptation \citep{ganin2016DAN} and deconfounding for text classification \citep{pryzant2018lexicon,elazar2018adv} assumes that the set of confounds is known a priori, and their values are given as part of the training data. This is an unrealistic setting that limits the applicability of such models in real world scenarios. In contrast, we introduce a new method, based on log-odds ratio with Dirichlet prior \citep{monroe2008logodds}, for identifying and representing latent confounds as probability distributions (\Sref{sec:confounds}).
Second, we propose a novel alternating learning procedure with multiple adversarial discriminators, inspired by adversarial learning \citep{goodfellow2014gans},  that demotes latent confounds and results in textual representations that are invariant to the confounds~(\Sref{sec:classifier}).


Note that these two proposals are task-independent and can be extended to a vast array of text classification tasks where confounding factors are not known a priori.
For concreteness, however, we evaluate our approach on the task of L1ID~(\Sref{sec:experiments}). 
We experiment with two different datasets: a small corpus of student written essays \citep{2017sharedtasknli} and a large and noisy dataset of Reddit posts \citep{rabinovich2018redditcorpus}. We show that classifiers trained on these datasets without any intervention learn spurious topical correlations that are not indicative of style, and that our proposed deconfounded classifiers alleviate this problem~(\Sref{sec:results}). We  present an analysis of the features discovered after demoting these confounds in~\Sref{sec:analysis}.

The main contributions of this work are: 
\begin{asparaenum}
\item We introduce a novel method for representing and identifying variables which are confounds in text classification tasks.
\item We propose a classification model and an algorithm aimed at learning textual representations that are invariant to the confounding variable. 
\item We introduce a novel approach to adversarial training with multiple adversaries, to alleviate the problem of drifting parameters 
during alternating classifier--adversary optimization.
\item Finally, we analyze some linguistic features that are not only predictive of the author's L1 but are also devoid of topical bias. 
\end{asparaenum}

\section{Motivation}
\label{sec:motivation-to-demote}
We study the general effect of \emph{topical} confounds in text classification. 
To motivate the need to demote them, we introduce as a case study the L1ID task, in which the goal is to predict the native language of a writer given their texts in L2. 

We begin with a subset of the L2-Reddit corpus \citep{rabinovich2018redditcorpus}, 
consistsing of Reddit posts by authors with~23 different L1s, most of them European languages. Some of the posts come from Europe-related forums (e.g. r/Europe, r/AskEurope, r/EuropeanCulture), whereas others are from unrelated forums. We view the latter as out-of-domain data and use them to evaluate the generalization of our models. 
We use a subset of this corpus, with only the~10 most frequent L1s, to guarantee a large enough balanced training set. We remove all the posts with fewer than 50 words and sample the dataset to obtain a balanced distribution of labels: from this balanced dataset, we randomly sample 20\%  of examples from each class and divide them equally to create development and test sets. 
In total, there are around 260,000 examples in the training set and 32,000 examples each in the development, the in-domain test set, and the out-of-domain test set. 

We trained a standard (non-adversarial) classifier, with a bidirectional LSTM encoder followed by two feedforward layers with a $\tanh$ activation function and a softmax in the final layer (full experimental details are given in \Sref{subsec:baseline-classifier}).  We refer to this model as \textsc{no-adv}. The results 
are shown in~\Tref{tab:topical-baseline}.
Notice the huge drop in accuracy on the out-of-domain data, which indicates that the model is learning topical features. 

To further verify this claim, we  used \emph{log-odds ratio with Dirichlet prior} \citep{monroe2008logodds}---a common way to identify words that are statistically overrepresented in a particular population compared to others---to identify the top-$K$ words that were most strongly associated with a specific L1 in the training set. (We refer the reader to \citep{monroe2008logodds} for the details about the algorithm.)
We experimented with $K \in \{20, 50, 100, 200\}$.
\Tref{tab:logodds} shows the top-$10$ words in each class; observe that almost all of these words are geographical (hence, topical) terms that have nothing to do with the L1. 

Next, we masked such topical words (by replacing them with a special token) and evaluate the trained classifier on masked test sets. 
Accuracy (\Tref{tab:topical-baseline}) degrades on both the in-domain and out-of-domain sets, even when only $20$ words are removed. The drop in accuracy with the out-of-domain dataset is smaller since these data do not include many instances where the presence of topical words would help in identifying the label. 
These experiments confirm our hypothesis that the baseline classifier is primarily learning topical correlations, and motivate the need for a deconfounded classification approach which we describe next.

\begin{table}[hbt]
\centering
\begin{tabular}{lrr}
 & \textbf{\begin{tabular}[c]{@{}l@{}}In-\\ Domain\end{tabular}} & \textbf{\begin{tabular}[c]{@{}l@{}}Out-of-\\ Domain\end{tabular}} \\ \hline
\textbf{\textsc{no-adv}} & 52.5 & 25.7 \\ 
\textbf{\textsc{+Mask Top-20}} & 32.8 & 21.0 \\
\textbf{\textsc{+Mask Top-50}} & 31.6 & 20.4 \\
\textbf{\textsc{+Mask Top-100}} & 30.1 & 19.7 \\
\textbf{\textsc{+Mask Top-200}} & 28.5  & 18.7 \\ 
\end{tabular}
\caption{Motivation: accuracy (\%) of L1ID on the L2-Reddit dataset.
\label{tab:topical-baseline}}
\end{table}

\begin{table*}[hbt]
\centering
{
\begin{tabular}{ll}
\textbf{English} & ireland irish british britain russia scotland england states american london brexit \\ \hline
\textbf{Finnish} & finland finnish finns helsinki swedish finn nordic sweden sauna nokia estonian \\ \hline
\textbf{French} & french france paris sarkozy macron fillon hollande gaulle hamon marine valls breton \\ \hline
\textbf{German} & german germany austria merkel refugees asylum germans bavaria austrian berlin also \\ \hline
\textbf{Greece} & greek greece greeks syriza macedonia athens turkey macedonians fyrom turkish ancient \\ \hline
\textbf{Dutch} & dutch netherlands amsterdam wilders rotterdam holland rutte belgium bike hague \\ \hline
\textbf{Polish} & poland polish poles warsaw lithuanian lithuania judges jews ukranians imho tusk  \\ \hline
\textbf{Romanian} & romania romanian romanians moldova bucharest hungarian hungarians transistria \\ \hline
\textbf{Spanish} & spain catalan spanish catalonia catalans madrid barcelona independence spaniards \\ \hline
\textbf{Swedish} & sweden swedish swedes stockholm swede malmo danish nordic denmark finland \\ 
\end{tabular}
}
\caption{Top words based on log-odds scores for each label in the L2-Reddit dataset.}
\label{tab:logodds}
\end{table*}

\section{Representing Confounds}
\label{sec:confounds}

Latent Dirichlet allocation (LDA; \citealp{bleiLDA2003}) is a probabilistic generative  model for discovering abstract topics that occur in a collection of documents. 
Under LDA, each document can be considered a mixture of a small (fixed) number of topics---each represented as a distribution over words---and each word's presence is assumed to be attributed to one of the document's topics. More precisely, LDA assigns each document a probability distribution over a fixed number of topics $K$. 

LDA topics are known to be poor features for classification  \citep{mcauliffe2008supervised}, indicating that they do not encode all the topical information. Moreover, they can  encode information which is not actually topical and can be a useful L1 marker. 
\ignore{There exist many supervised versions of LDA as well \citep{mcauliffe2008supervised,labeledLDA} but they also are likely to encode more information than the actual confounds.} 
Motivated by our case study (\Sref{sec:motivation-to-demote}), we propose a novel method to represent topic distributions, based on log-odds scores \citep{monroe2008logodds}, and compare it to LDA as a baseline. 

For each class label $y$ and each word type $w$, we calculate a log-odds score $\mathit{lo}(w, y) \in \mathbb{R}$. The higher this score, the stronger the association between the class and the word. As we saw in \Sref{sec:motivation-to-demote}, the highest scored words are mostly topical and hence constitute superficial features which we want the classification model to ``unlearn.'' We therefore define a distribution which assigns high probability to a document containing these high scoring words. For a label $y \in \mathcal{Y}$ and an input document $x = \langle w_1, \ldots, w_n\rangle$, we define $p(y \ | \ x )$:
\[
    p(y \ | \ x) \propto p (y) \cdot p(x \ | \ y)  
    = p (y) \cdot \prod_{i=1}^n p (w_i \ | \ y) 
\]
The above expansion assumes a bag of words representation. When the dataset is balanced, $p(y)$ is equal for each label and can be omitted. Finally, we define
    $p(w_i \ | \ y) \propto \sigma(lo(w_i, y))$,
where $\sigma(.)$ is the sigmoid function, which squashes the log-odds scores (whose values are in $\mathbb{R}$) to the range $[0, 1]$. We normalize the sigmoid values over the vocabulary to convert them to a probability distribution. In this distribution, the number of ``topics'' equals the number of labels, $m$. 

\section{Deconfounded Text Classification}
\label{sec:classifier}

\begin{figure*}[hbt]
  \begin{subfigure}[b]{0.48\textwidth}
    \includegraphics[width=\textwidth]{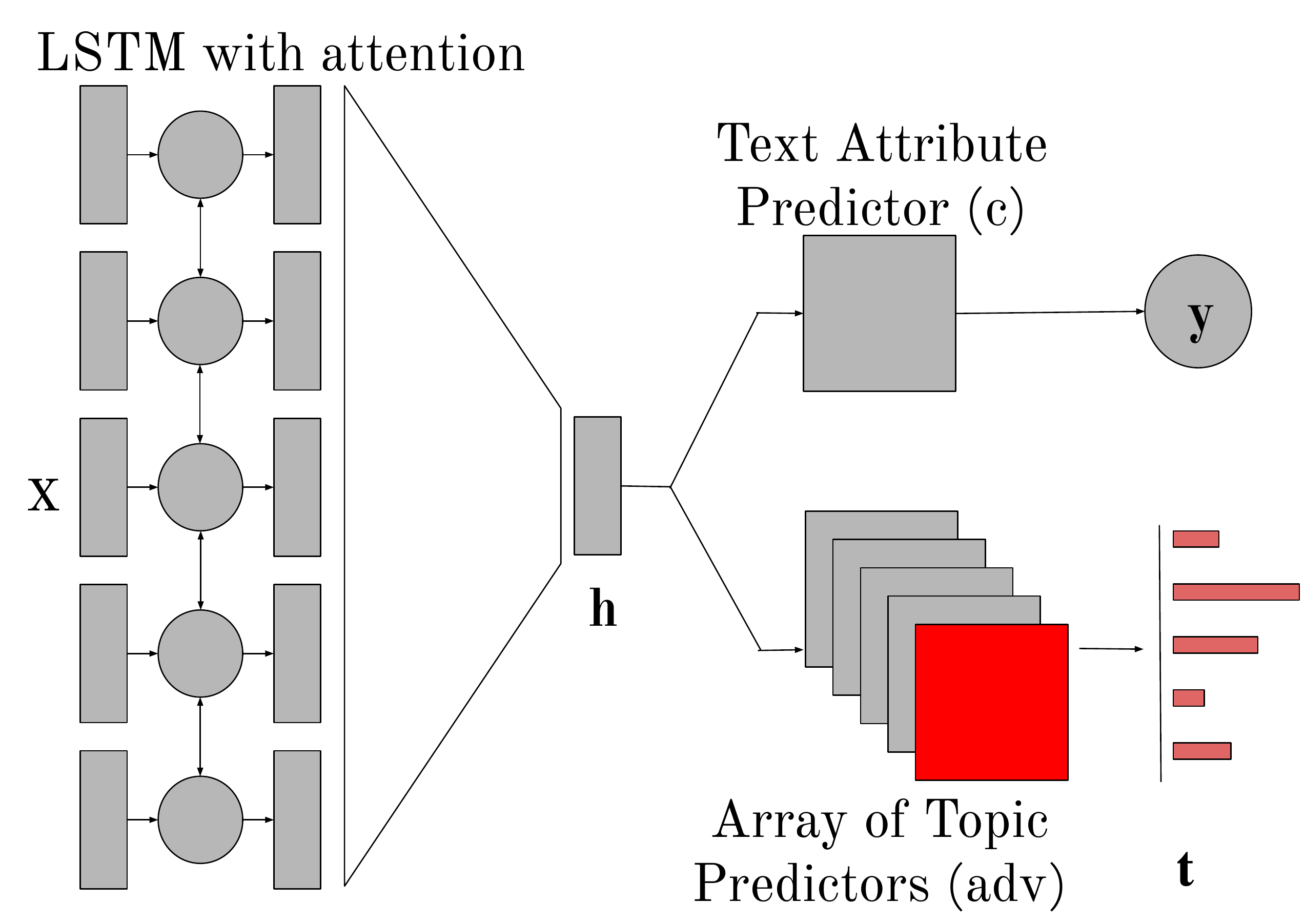}
    \caption{Weights of the LSTM and of the discriminator are fixed. A new topic predictor is trained by minimizing the cross entropy of the output and the distribution of the input document over latent topics as described in \Sref{sec:confounds}.}
    \label{fig:topic-training}
  \end{subfigure}
  \hfill
  \begin{subfigure}[b]{0.48\textwidth}
    \includegraphics[width=\textwidth]{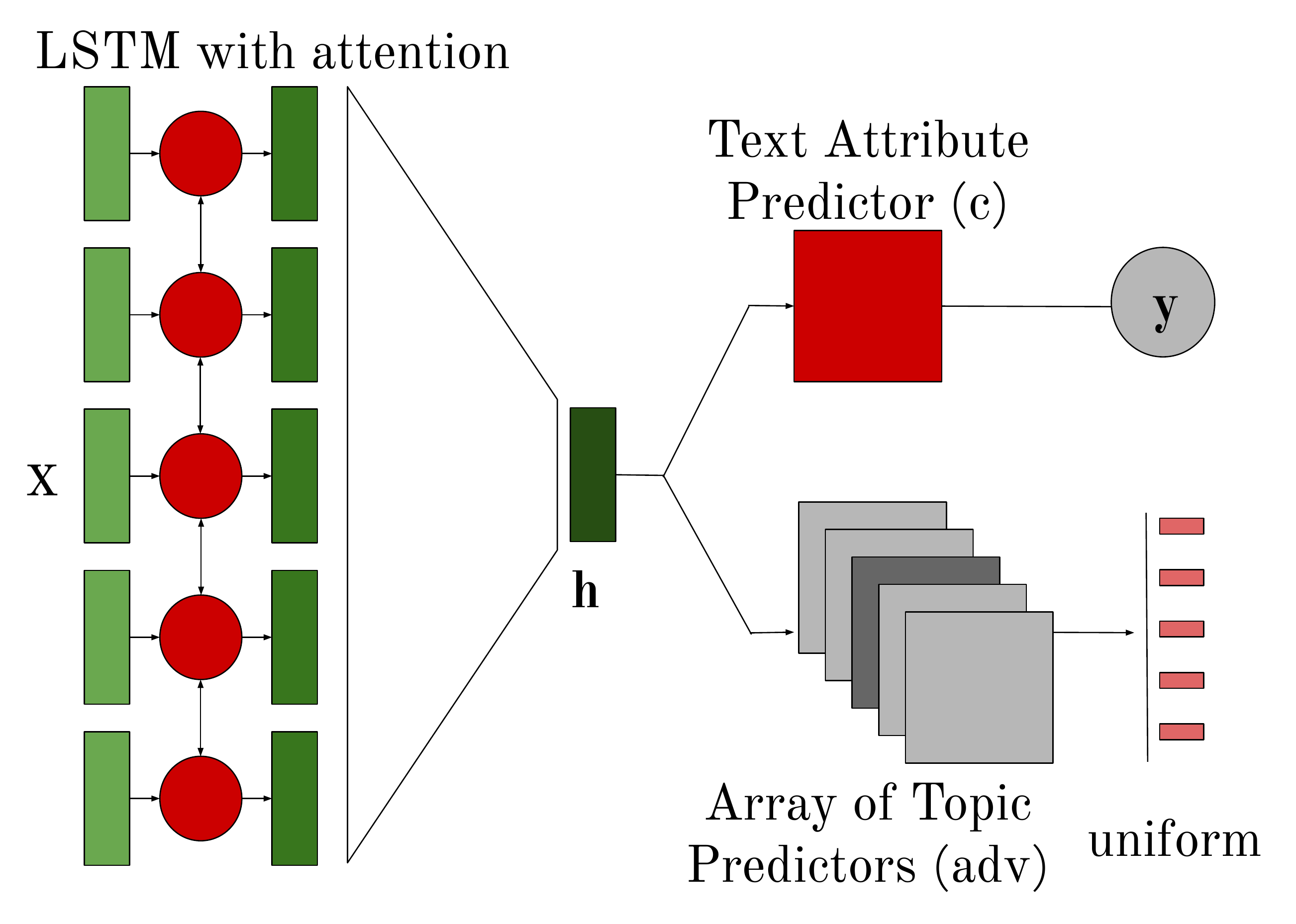}
    \caption{Weights of all the topic predictors are fixed, but the encoder is trained. The model is jointly minimizing the cross-entropy 
    of the classifier and encouraging  
    the topic predictor toward uniformity.}
    \label{fig:topic-forgetting}
  \end{subfigure}
  \caption{We alternate between training the topic predictor (left; \eqref{eq:topic-loss}) and the deconfounded classifier/encoder (right;  \eqref{eq:classifier-loss}). Pretraining is not shown in the figure.} 
  \label{fig:model}
\end{figure*}


We now formalize the task setup and the classification model.
We are given $N$ labeled documents in the training set $\{(x_1, y_1), (x_2, y_2),$ $ \ldots, (x_N, y_N)\}$, where  $x_i$ is a document  with  label $y_i \in \mathcal{Y}$, where $m = | \mathcal{Y} |$ is the number of labels. 
For each document $x_i$, we represent latent (topical) confounds---domain-specific and superficial document features---as a $K$-dimensional multinomial distribution 
$t_i \in \{(t_1, \ldots, t_K) \ | \ \sum_{j=1}^K t_j = 1 \}$.
In our task, the confounds are \emph{topics}, so that each $t_j$ represents the proportion of document $i$ associated with topic $j$ but these topics are not given a priori. 
In this work, the number of topics $K$, equals $m$, but the methods presented in this work are valid for any number of topics.

Our goal is to train a classifier $f$, parameterized by $\theta$, which learns to accurately predict the target label, while ignoring superficial topical correlations present in the training set. That is, for a text $x$ we wish to predict $\hat{y} = f_\theta(x)$ which doesn't encode any information about $t$. Following \citet{ganin2016DAN}, \citet{pryzant2018lexicon}, and \citet{elazar2018adv}, we input $x$ to an encoder neural network $h(x; \theta_h)$ to obtain a hidden representation $\mathbf{h}_x$ (see \Fref{fig:model}), followed by two feedforward networks:
\begin{inparaenum}[(1)]
\item $c(h(x); \theta_c)$ to predict the label $y$; and 
\item an adversary network $\mathrm{adv}(h(x); \theta_a)$ to predict the topics. 
\end{inparaenum} Departing from prior work which used predefined binary 
confounds, our adversary predicts the topic distribution $t$. If $\mathbf{h}_x$ does not encode any information to predict $t$, then $c(h(x))$ will not depend on $t$. Concretely, we want to optimize the following quantity: 
\begin{align*}
    \min_{c, h}   \frac{1}{N} \sum_{i=1}^N & \mathrm{CE}(c(h(x_i)), y_i) \\ + 
    & \mathrm{CE}(\mathrm{adv}^\ast_h (h(x_i) ), \mathbb{U}_K ) \nonumber
\end{align*}
where $\mathrm{CE}$ denotes cross-entropy loss, and
\begin{align*}
\mathrm{adv}^\ast_h = \arg \min_{\mathrm{adv}}  \frac{1}{N} \sum_{i=1}^N \mathrm{CE}(\mathrm{adv}(h(x_i)), t_i),
\end{align*}
and $\mathbb{U}_K = (\frac{1}{K}, \ldots, \frac{1}{K})$.
This objective seeks a representation $\mathbf{h}_x$ which is maximally predictive of the class label but not of the topical distribution (ideally, it should output a uniform topic distribution for every input).


\subsection{Learning Schedule: Alternating Optimization of Classifier and Adversary}

In practice, this optimization is done in an alternating fashion by minimizing the following two quantities:
\begin{align}
     & \begin{aligned}\min_{\mathrm{adv}} \frac{1}{N} \sum_{i=1}^N \mathrm{CE}(\mathrm{adv}(h(x_i)), t_i) \label{eq:topic-loss} 
     \end{aligned} \\
     & \begin{aligned}
        \min_{c, h} \frac{1}{N} \sum_{i=1}^N &\mathrm{CE}(c(h(x_i)), y_i) \\ 
        & +\mathrm{CE}(\mathrm{adv}(h(x_i)), \mathbb{U}_K)  \label{eq:classifier-loss}
     \end{aligned} 
\end{align}
The training schedule is critical in adversarial setups where the loss has two competing terms \citep{mescheder2018training,arjovsky2017gan,roth2017stabilizing}; here, these terms minimize classification loss while maximizing the topic prediction loss.  
Algorithm~\ref{alg} details our proposed alternating learning procedure.

 \begin{algorithm}[ht]
\KwResult{ $\theta_h, \theta_c, \theta_{a_1}, \ldots, \theta_{a_T}$ }
 Randomly initialize $\theta_h, \theta_c$\;
 \While{not converged}{
    Sample a minibatch of $b$ training samples\;
    
    Update $\theta_h$ and $\theta_c$ using gradients with \\ respect to $\frac{1}{b} \sum_{i=1}^b \mathrm{CE} (c(h(x_i)), y_i)$.\;
 }
 $j=1$\;
 
 \For{number of training iterations $T$}{
    Randomly initialize $\theta_{a_j}$\;}
    
    \For{$t$ steps}{
    Sample a minibatch of $b$ training samples\;
    
    Fix $\theta_h$ and $\theta_c$, update $\theta_{a_j}$ using gradients with respect to $\frac{1}{b}\sum_{i=1}^b \mathrm{CE}(\mathrm{adv}_{\theta_{a_j}}(h(x_i)), t_i)$\;
    }
  \For{$c$ steps}{ 
    Sample a minibatch of $b$ training samples\;
    
    Fix $\theta_{a_u}$ for $u \in_R \{1, \ldots, j\}$ and update $\theta_c$ and $\theta_{h}$ using gradients with respect to
    $\frac{1}{b} \sum_{i=1}^b \mathrm{CE}(c(h(x_i)), y_i) +  \mathrm{CE}(\mathrm{adv}_{\theta_{a_u}}(h(x_i)), \mathbb{U}_K)$\; 
    } 
    $j \longleftarrow j + 1$\;
 \caption{Alternating optimization of classifier and adversary.}
 \label{alg}
\end{algorithm}
Inspired by generative adversarial networks (GANs;  \citealp{goodfellow2014gans}), the training procedure  alternates between training the classifier and the adversary (see \Fref{fig:model}). First (\emph{pretraining}), we train the encoder along with the classifier using only classification loss,  until convergence. After pretraining, $\mathbf{h}_x$ has encoded topical information which it uses for classification (as shown in our analysis in \Sref{sec:motivation-to-demote}). Now, we train only $\mathrm{adv}(h(x))$ to (accurately) 
predict $t$, keeping the parameters of $h(.)$ fixed. Once $\mathrm{adv}(.)$ is trained, it should be able to successfully extract a topic distribution from $\mathbf{h}_x$ (topic training, see \Fref{fig:topic-training}). The goal now is to modify $\mathbf{h}_x$ in such a way that $\mathrm{adv}(\mathbf{h}_x)$ produces a uniform distribution (that is, fooling the adversary; similar to fooling the discriminator in GANs). We do that by keeping the weights of $\mathrm{adv}(.)$ fixed, and training the network to produce the class label and a uniform topic distribution (\emph{topic forgetting}, see \Fref{fig:topic-forgetting}). We then repeat this procedure for a fixed number of steps which was tuned using the validation set.

\subsection{Multiple Adversaries} 
In our experiments, we observe that after every ``topic forgetting'' stage, $\mathrm{adv}(.)$ does end up producing a uniform distribution, but in the next ``topic training'' phase, $\mathrm{adv}(.)$ is able to reproduce the topical distribution accurately. This is because, during ``topic forgetting,'' the classifier does not really forget the topics in $\mathbf{h}_x$; it just encodes them in a different way.\footnote{We observe this in our analysis where the most salient features encoded after pretraining and topic forgetting phrase are the same.}  This is a general problem in setups with alternating classifier-adversary optimization. To solve this issue, we propose using \emph{multiple adversaries}, inspired by the ``experience replay'' approach used in reinforcement learning \cite{o2010play,mnih2015human}. During the  $i$th ``topic training'' phase, we train a new adversary $\mathrm{adv}_i$ (with parameters $\theta_{a_i}$ instead of retraining only one adversary over and over again. In the next ``topic forgetting'' phase, at each training step we pick $\mathrm{adv}_j$ at random from the pool of previously learned adversaries, $j \in_R \{1, \ldots, i\}$. By using multiple adversaries, we make it difficult for the classifier to encode topical information anywhere. 



\section{Experimental Setup}
\label{sec:experiments}

\subsection{Datasets}

We evaluate our topical confound demotion method on the L1ID task. We show experiments with two datasets where  L2 is English: the L2-Reddit dataset  described in \Sref{sec:motivation-to-demote}, and TOEFL17, a collection of essays authored by non-native English speakers who apply for academic studies in the~US \citep{2017sharedtasknli}. 
This corpus reflects eleven L1s: Arabic, Chinese, French, German, Hindi, Italian,
Japanese, Korean, Spanish, Telugu, and Turkish. 
The training data include 11,000 authors (1,000 per L1) and the development set has 1,100 essays per L1. We evaluate on the development set. Each essay is also marked with a prompt~ID which was given to the authors to write the essay. There are 8 prompts in total, based on which we construct 8 versions of train and test set. In each version, we remove essays marked with one of the prompts from both the train and the development sets, and consider the removed essays from the development set an ``out-of-domain'' test set. We refer to the version where prompt ``P$K$'' is out-of-domain as ``--P$K$'' in the results (\Tref{results:toefl-final}), $K \in \{0, \ldots, 7\}$. 

\subsection{Implementation Details}
\label{subsec:baseline-classifier}
We tokenized and lowercased all the text using  \href{https://spacy.io/}{spaCy}. Limiting our vocabulary to the most frequent 30,000 words in the training data, we replaced all  out-of-vocabulary words with ``UNK.'' We encoded each word using a word embedding layer (initialized at random and learned) and passed these embeddings to a bidirectional LSTM encoder (one layer for each direction) with attention ($h(x)$; \citealp{pryzant2018lexicon}). Each LSTM layer had a hidden dimension of 128. We used two layered feed forward networks with a $\tanh$ activation function in the middle layer (of size 256), followed by a softmax in the final layer, as $c(.)$ and $\mathrm{adv}(.)$.

\subsection{Baselines}
We consider several baselines that are intended to capture the stylistic features of the texts, explicitly avoiding  content.

\paragraph{Linear classifier with content-independent features (LR)} Replicating \citet{goldin2018l1id}, we trained a logistic regression classifier with three types of features: function words, POS trigrams, and sentence length, all of which are reflective of the style of writing. We deliberately avoided using content features (e.g., word frequencies). 

\paragraph{Classification with no adversary on masked texts (\textsc{lo-top-$K$})}
We mask the top-$K$ words (based on log-odds scores) in \emph{both} the train and the test sets (as in \Sref{sec:motivation-to-demote}); we train the classification model again without training $\mathrm{adv}(.)$. After masking the top words, we expect patterns of writing style (and, therefore, L1) to become more apparent.

\paragraph{Adversarial training with gradient reversal (\textsc{gr-lo})} 
A common method of learning a confound-invariant representations is to use a gradient reversal layer \citep{beutel2017data,ganin2016DAN, pryzant2018lexicon,elazar2018adv}. The output of the encoder, $\mathbf{h}_x$, is passed through this layer before applying $\mathrm{adv}(.)$. This training setup usually proves too difficult to optimize, and often results in poor performance. That is, even if the performance of $\mathrm{adv}(.)$ is weak, $\mathbf{h}_x$ still ends up leaking information about the confound \cite{lample2018multipleattribute,elazar2018adv}. In the forward pass, this layer acts as identity whereas in the backward pass it multiplies the gradient values by $-\lambda$,  essentially reversing the gradients before they go into the encoder. $\lambda$ controls the intensity of the reversal (we used $\lambda=0.2$). 

\paragraph{LDA topics as confounds (\textsc{alt-lda})} We trained LDA on the training set and for each example in the training set, generated a probability distribution (over $50$ topics), and used it as topical confound with our proposed learning setup, alternating classifier-adversary training. 

\section{Results}
\label{sec:results}

\subsection{TOEFL17 Dataset}
We begin with experiments on the TOEFL17 dataset, where predicting L1 is an easier task due to the lower proficiency  of the authors.
\Tref{results:toefl-final} reports the accuracy of our proposed model, denoted  \textbf{\textsc{alt-lo}}, compared to the logistic regression baseline (\textbf{\textsc{lr}}), and two adversarial baselines: one demotes latent log-odds-based topics via gradient reversal  (\textbf{\textsc{gr-lo}}), and another uses our proposed novel learning procedure but demotes baseline LDA topics (\textbf{\textsc{alt-lda}}). 
We report both in-domain accuracy and out-of-domain results;
the latter is obtained by averaging the accuracy of each set ``--P$K$'' over $K \in \{0, \ldots, 7\}$. 

\begin{table}[hbt]
\centering
\begin{tabular}{lrr}
 & \textbf{\begin{tabular}[c]{@{}l@{}}In-\\ Domain\end{tabular}} & \textbf{\begin{tabular}[c]{@{}l@{}}Out-of-\\ Domain\end{tabular}} \\ \hline
\textbf{\textsc{lr}} & 55.3 & 50.9 \\
\textbf{\textsc{gr-lo}} & 12.7 & 13.6 \\
\textbf{\textsc{alt-lda}} & 59.1 & 50.1 \\ \hline
\textbf{\textsc{alt-lo}} & \textbf{61.9} & \textbf{60.4} \\ 
\end{tabular}
\caption{Classification accuracy with topic-demoting methods, TOEFL dataset.}
\label{results:toefl-final}
\end{table}

Our model strongly outperforms all baselines that demote confounds, in both classification setups.  
We observe in our experiments that gradient reversal is especially unstable and hyperparameter sensitive: it has been shown to work well with categorical confounds like domain type or binary gender, but in demoting continuous outputs like a topic distribution, we observe it is not effective. The proposed alternating training with multiple discriminators obtains better results, and replacing LDA with log-odds-based topics also improves both in-domain and (much more substantially)  
out-of-domain predictions, confirming the effectiveness of our proposed innovations.

A vanilla classifier without demoting confounds (denoted in \Sref{sec:motivation-to-demote} as \textsc{no-adv}) yields in-domain and out-of-domain accuracies of 62.0 and 58.3, respectively. We would  expect that the better generalization power of our proposed model would come at a price of lower accuracy in-domain. Our goal is to capture the true signals of L1, rather than superficial patterns that are more frequent in the data and artificially boost the performance in \textsc{no-adv} settings. This is indeed what we observe.

For example, the text ``\ldots i agree with you on the prolonged war if the plc heartland (poland proper) was not as rich as it was i dont really see how we would been \ldots'' in the dataset is labeled as ``Polish'' instead of the gold label ``Swedish'' by the \textsc{no-adv} classifier, likely because of the mention of the term ``poland'', but the \textsc{adv-lo} model predicts it correctly since it likely picks on other features that indicate non-fluency,  like 
``we would been''.
Such naive classification errors become especially costly in making predictions about people's demographic attributes: ethnicity, which often correlates with L1, but also gender, race, religion, and others  \cite{hardt2016equality,beutel2017data}.

\subsection{L2-Reddit Dataset}
Next, we experiment with L2-Reddit, a larger and more challenging dataset (since many speakers in the dataset are highly fluent, and the signal of their native language is weaker).  
The performance of the simple baselines on  this dataset is shown in \Tref{results:reddit10}. The accuracy of the linear classifier is poor (compared to \Tref{tab:topical-baseline}), perhaps because it fails to capture some contextual features learned by the neural network models. 
With \textsc{lo-top-20}, the performance on both test sets improves. It slightly degrades when more words are removed, perhaps because some words indicative of \textsc{L1} are also removed. 

\begin{table}[hbt]
\centering
\begin{tabular}{lrr}
 & \textbf{\begin{tabular}[c]{@{}l@{}}In-\\ Domain\end{tabular}} & \textbf{\begin{tabular}[c]{@{}l@{}}Out-of-\\ Domain\end{tabular}} \\ \hline
\textbf{\textsc{lr}} & 21.2 & 18.5 \\ 
\textbf{\textsc{lo-top-20}} & 38.7 & 21.9 \\
\textbf{\textsc{lo-top-50}} & 36.4 & 21.4 \\
\textbf{\textsc{lo-top-100}} & 35.8 & 21.2 \\
\textbf{\textsc{lo-top-200}} & 34.7 & 20.8 \\ 
\end{tabular}
\caption{Baseline classification accuracy on L2-Reddit.}
\label{results:reddit10}
\end{table}

Finally, we evaluate the impact of our novel training procedure and the quality of our proposed topical confound identification method.  We compare our proposed solution, denoted \textsc{alt-lo}, with two alternatives, as before, one with a different learning setup (\textsc{gr-lo}) and one with a different confound representation (\textsc{alt-lda}).
\Tref{results:reddit-final} summarizes the results: our proposed learning procedure \textsc{alt-lo} performs better than both the alternatives. Unsurprisingly, the model trained with gradient reversal (\textsc{gr-lo}) performs particularly poorly; this was our primary motivation to explore better learning techniques.
\begin{table}[hbt]
\centering
\begin{tabular}{lrr}
 & \textbf{\begin{tabular}[c]{@{}l@{}}In-\\ Domain\end{tabular}} & \textbf{\begin{tabular}[c]{@{}l@{}}Out-of-\\ Domain\end{tabular}} \\ \hline
\textbf{\textsc{gr-lo}} & 22.5 & 15.7 \\
\textbf{\textsc{alt-lda}} & 46.2 & 21.9 \\ \hline
\textbf{\textsc{alt-lo}} & \textbf{48.8} & \textbf{22.9} \\ 
\end{tabular}
\caption{Classification accuracy with topic-demoting methods, L2-Reddit dataset.}
\label{results:reddit-final}
\end{table}

To further confirm that the  \textsc{alt-lo} model is not learning topical features, we repeat the experiment presented in \Tref{tab:topical-baseline}---masking the top $K$ topical words (based on log-odds scores) from the test sets, but not retraining the models---now, with our proposed model  \textsc{alt-lo}. \Tref{tab:removetopk} shows that 
in contrast to standard models that do not demote topical confounds (as in \Tref{tab:topical-baseline}),  
there is less degradation in the performance of \textsc{alt-lo}.
We conjecture that our model is stable to demoting topics because it learns relevant stylistic features, rather than spurious correlations.

\begin{table}[hbt]
\centering
\begin{tabular}{lrr}
 & \textbf{\begin{tabular}[c]{@{}l@{}}In-\\ Domain\end{tabular}} & \textbf{\begin{tabular}[c]{@{}l@{}}Out-of-\\ Domain\end{tabular}} \\ \hline
\textbf{\textsc{alt-lo}} & 48.8 & 22.9 \\ 
\textbf{+\textsc{Mask Top-20}} & 38.7 & 21.6 \\
\textbf{+\textsc{Mask Top-50}} & 36.2 & 21.5 \\
\textbf{+\textsc{Mask Top-100}} & 33.5 & 21.2 \\
\textbf{+\textsc{Mask Top-200}} & 31.9 & 20.4 \\ 
\end{tabular}
\caption{Accuracy on the L2-Reddit dataset; the proposed model (\textsc{alt-lo}) with different settings of the test sets.}
\label{tab:removetopk}
\end{table}

\ignore{
\sw{I feel that much more explanation is needed here to drive home the conclusion that our proposed model is better.} \nascomment{I agree.  the current text shows a lot of numbers and heads off a lot of anticipated concerns, but doesn't explain to the reader what we're trying to answer with each of the various experiments.  One thing I'm missing in particular is how we are assessing whether a model is picking up on the wrong cues as opposed to the right ones.} \sk{I have tried to add some more explanation in the previous two paragraphs addressing your concerns. How does it look now? Noah: We assess that the model is picking up on wrong cues by removing those cues from the text and looking at the accuracy (table 2 and 6)}
}

\section{Analysis}
\label{sec:analysis}

We present an analysis of what the models are learning, based on words they attend to for classification. We focus on the L2-Reddit dataset. 

Following \citet{pryzant2018lexicon}, we generated a lexicon of most attended words by (1) running the model on the test set and saving the attention score for each word; and (2) for each word, computing its average attentional score and selecting the top-$k$ words based on this score.

What emerges from this lexicon (\Tref{tab:lexicon}) is a dramatic difference between the top indicative words in the various models. Whereas in the baseline model \emph{all} the most indicative words are proper nouns, the \textsc{alt-lo} model highlights exclusively function words. The proper nouns in the baseline model are all geographical terms directly associated with the L1s reflected in the L2-Reddit dataset: they are easy giveaways of the authors' L1s, but they are meaningless linguistically. In contrast, the function words highlighted in the \textsc {alt-lo} model are mostly prepositions and determiners; it is well known that nonnative speakers are challenged by the use of prepositions (in any L2, English included). The distribution of determiners is also a challenge for nonnatives, and the correct usage of \textsl{the} in particular is quite hard for learners to master. These challenges are evident from the most indicative words of our model. Observe also that the \textsc{lo-top-$50$} model is somewhere in the middle: it includes some proper nouns (including geographical terms such as \textsl{eu} or \textsl{us}) but also several function words. A more detailed analysis of these observations is left for future work.

Recently, there has been a debate on whether attention can be used to explain model decisions \citep{serrano-attention,attnnotexplain,attnnonottexplain}, 
we thus present additional analysis of our proposed method based on saliency maps \citep{saliency}. 
Saliency maps have been shown to better capture word alignment than attention probabilities in neural machine translation. This method is based on computing the gradient of the probability of the predicted label with respect to each word in the input text and normalizing the gradient to obtain probabilities. 
We use saliency maps to generate lexicons similar to the ones generated using attention. As shown in \tref{tab:lexicon-saliency}, the top indicative words for baseline and \textsc{lo-top-50} follow a similar pattern as the ones obtained with attention scores. In line with results in \Tref{tab:lexicon}, salient words for \textsc{alt-lo} are determiners and prepositions.  
However, saliency maps also reveal that our proposed approach still attends to some geographical terms that were not demoted by our classifier. 

\begin{table*}[hbt]
\centering
\begin{tabular}{ll}
\hline
\textbf{\textsc{no-adv}} & \begin{tabular}[c]{@{}l@{}}sweden france greece finland poland spain greek germany french eu\\ romania polish dutch german spanish swedish netherlands finnish\end{tabular} \\ \hline
\textbf{\textsc{lo-top-50}} & \begin{tabular}[c]{@{}l@{}}eu 's 're 'm ' \& uk us because 've am its nt english these usa n’t \\ here 'll especially correct pis de within\end{tabular} \\ \hline
\textbf{\textsc{alt-lo}} & \begin{tabular}[c]{@{}l@{}}the in to of that a i is and 't as from with by ? on but \& they\\ are about at because like was would have you\end{tabular} \\ \hline
\end{tabular}
\caption{
The highest scoring words in lexicons generated using attention scores.
\label{tab:lexicon}}
\end{table*}


\section{Related Work}
\label{sec:related}
\paragraph{Controlling for confounds in text}
Controlling for confounds is an active field of research, especially in the medical domain, where the common solution is to do random trials or propensity score matching \citep{propensity}. \citet{Paul2017FeatureSA} tackled the problem of learning causal associations between word features and class labels using propensity matching for the task of sentiment analysis. 
This method is not scalable to large text datasets as it involves training a logistic regression model for every word type. \citet{tantwitter} built models to estimate the number of retweets of Twitter messages and addressed confounding factors by matching tweets
of the same author and topic. \citet{robustclassify2018} proposed a statistical technique called Pearl's back-door adjustment for text classification \citep{Pearl:2009:CMR:1642718}. All these works focused on a bag-of-words model with lexical features only.  

\paragraph{Adversarial training in text} Much recent work focuses on learning textual representations that are invariant to selective properties of the text. This work used domain adaptation and transfer learning \citep{ganin2016DAN, tzeng2014DeepDC, xie2018nips}, either to remove sensitive attributes such as  demographic information 
\citep{li2018privacy, elazar2018adv, beutel2017data, coavoux-etal-2018-privacy}, or to understand costumer 
behavior for social science applications \citep{pryzant2018lexicon}. Most of the work in this area, however, focuses on cases where these confounds are known in advance and their values are given along with the training data. Our presented approach is most closely related to~\citet{coavoux-etal-2018-privacy} who proposed an alternating optimization method to learn privacy-preserving text representations. This work focuses on demoting binary-valued attributes by maximizing the likelihood of erroneous label using a single adversary network, which we have shown to be inadequate in our experiments.  In constrast, we propose a more general method focusing on multinomial distributions which we push towards a  uniform distribution with the help of multiple adversaries. 

\paragraph{Native language identification}
The L1ID task was introduced by \citet{koppel2005automatically}, who worked on the International Corpus of Learner English \citep{icle}.
The same experimental setup was adopted by several other authors \citep{tsur2007using,Wong-Dras:2009:ALTA2009,wong-dras:2011:EMNLP}. Since the release of nonnative \emph{TOEFL} essays by the Educational Testing Service \citep{toefl}, the task gained popularity and this dataset has been used for two L1ID Shared Tasks \citep{tetreault-blanchard-cahill:2013:BEA,2017sharedtasknli}.

\citet{DBLP:journals/corr/MalmasiD17} report that the state of the art 
is a linear classifier with character $n$-grams and  lexical and morphosyntactic features. 
%

\begin{table*}[hbt]
\centering
\begin{tabular}{ll}
\hline
\textbf{\textsc{no-adv}} & \begin{tabular}[c]{@{}l@{}}poland  greek romania greece france spain french sweden finland \\ polish dutch spanish netherlands finnish  german\end{tabular} \\ \hline
\textbf{\textsc{lo-top-50}} & \begin{tabular}[c]{@{}l@{}}on 're even 'd up less things 'll doesn living majority \\sense talk level 've rights took number north\end{tabular} \\ \hline
\textbf{\textsc{alt-lo}} & \begin{tabular}[c]{@{}l@{}}the of to i a in greece romania france finland that for is french \& you \\'t finnish\end{tabular} \\ \hline
\end{tabular}
\caption{
The highest scoring words in lexicons generated using saliency maps.
\label{tab:lexicon-saliency}}
\end{table*}

The best accuracy under cross-validation on the TOEFL17 dataset, which includes~11 native languages (with a rather diverse distribution of language families), was 85.2\%.

The above works all identify the L1 of \emph{learners}. Identifying the native language of advanced, fluent speakers is a much harder task. \citet{goldin2018l1id} addressed this task, using the L2-Reddit dataset with as many as 23 different L1s, all of them European and many which are typologically close, which makes the task even harder. They experimented with a variety of features, using logistic regression as the classifier, and achieved results as high as 69\% accuracy with cross-validation; however, when testing their classifier outside the domain it was trained on (Reddit forums focusing on European issues), accuracy dropped to~36\%.

\section{Conclusion}
We introduced a method to represent unknown confounds in text classification using topic models and log-odds scores, and a new general method with alternating optimization to learn textual representations which are invariant of confounds. We evaluated the proposed solution on the task of native language identification, and showed that it learns to make predictions using stylistic features, rather than focus on topical information. 

The learning procedure we presented is general and applicable to other tasks that require learning invariant representations with respect to some attribute of text (some of which are discussed in \Sref{sec:related}). 
We plan to evaluate our proposed solution on other tasks where topics can be latent confounds, like predicting gender bias \citep{voigt2018rtgender}. We leave this exploration for future work.


\section*{Acknowledgments}
The authors acknowledge helpful input from the anonymous reviewers.
This work was supported in part by NSF grants IIS-1812327 and IIS-1813153, by grant no.~2017699 from the United States-Israel Binational Science Foundation (BSF), and by grant no.~LU~856/13-1 from the Deutsche Forschungsgemeinschaft.  Finally, the authors also thank Anjalie Field, Biswajit Paria, Ella Rabinovich, and Gili Goldin for helpful discussions.



\bibliography{emnlp-ijcnlp-2019}
\bibliographystyle{acl_natbib}

\end{document}